# Design of a Low-cost Miniature Robot to Assist the COVID-19 Nasopharyngeal Swab Sampling

Shuangyi Wang, Kehao Wang, Hongbin Liu and Zengguang Hou

*Abstract*—Nasopharyngeal (NP) swab sampling is an effective approach for the diagnosis of coronavirus disease 2019 (COVID-19). Medical staffs carrying out the task of collecting NP specimens are in close contact with the suspected patient, thereby posing a high risk of cross-infection. We propose a low-cost miniature robot that can be easily assembled and remotely controlled. The system includes an active end-effector, a passive positioning arm, and a detachable swab gripper with integrated force sensing capability. The cost of the materials for building this robot is 55 USD and the total weight of the functional part is 0.23kg. The design of the force sensing swab gripper was justified using Finite Element (FE) modeling and the performances of the robot were validated with a simulation phantom and three pig noses. FE analysis indicated a 0.5mm magnitude displacement of the gripper's sensing beam, which meets the ideal detecting range of the optoelectronic sensor. Studies on both the phantom and the pig nose demonstrated the successful operation of the robot during the collection task. The average forces were found to be 0.35N and 0.85N, respectively. It is concluded that the proposed robot is promising and could be further developed to be used in vivo.

## I. Introduction

The outbreak of novel coronavirus pneumonia (NCP) caused by coronavirus disease 2019 (COVID-19) has spread rapidly globally. Collection of specimens from the surface of the respiratory mucosa with nasopharyngeal (NP) or Oropharyngeal (OP) swabs are treated as effective ways for the diagnosis and screening [1]. Several recent studies have indicated that OP swabs are less effective than NP swabs in detecting the COVID-19 virus [2, 3] and concluded that the use of NP may be more suitable, although a study also highlighted the data should be viewed with cautions [4]. According to the US CDC, both NP and OP should be performed by a healthcare professional, while other possible approaches, e.g., nasal mid-turbinate (NMT) swab (also known as deep nasal swab) and anterior nares (nasal swab) specimen, could be supervised onsite self-collection and home self-collection [5]. During the conventional manually controlled swab sampling, medical staffs are unavoidably in close contact with the suspected patient, which poses a high risk of cross-infection. Medical workers' operating skills and psychological states may also affect the accuracy and quality of swab collection results.

As the global fight against COVID-19 may last for a long period of time with tens of thousands nasal swab samplings performed worldwide each day, robotic-assisted NP and OP swabbing with remote operation capability may reduce the risk of infection and free up staffs for other tasks. Comparing with human, the robot can be more thoroughly disinfected and those parts that are in close contact with patients can be disassembled and replaced. The value of using robots in fighting against COVID-19 has been highlighted in [6] and a well-designed robot for OP collection was recently developed by the Guangzhou Institute of Respiratory Health and the Shenyang Institute of Automation under the Chinese Academy of Sciences [7]. The robot consists of a snaking-shape robot arm for motion control, a binocular endoscope for visualization, a wireless transmission device and a human-computer interaction terminal for remote control. The system has been successfully tested in vivo and proved effective in clinical trials.

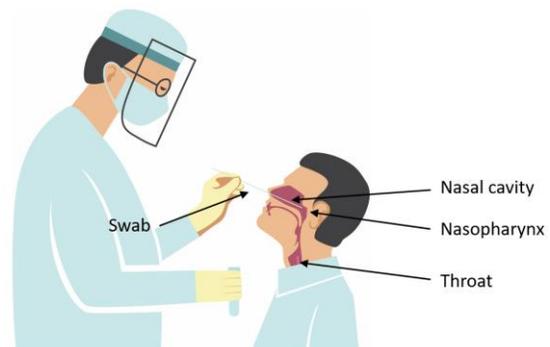

Fig. 1. Diagram showing the nasopharyngeal (NP) swab sampling procedure.

In the present study, we aim to explore a different strategy and develop a low-cost, easy-to-assemble robot with small footprint to assist NP swab collections (Figure 1). The NP swab collections involve inserting a specifically manufactured swab into a patient's nasal cavity [8]. The head of the patient is expected to be tilted back (at approximate 70°) so the nasal passage becomes straight and accessible. The swab is then inserted through the nostril parallel to the palate, all the way to the nasopharynx. After left in place for several seconds for secretion absorptions, the swab is then rotationally retracted from the nasal cavity slowly. In this paper, we aim to present

This work was funded and supported by the Chinese Academy of Sciences, Institute of Automation (CASIA) under the project No. Y9S8FZ0101.

S. Wang and Z. Hou are with the State Key Laboratory of Management and Control of Complex Systems, Chinese Academy of Sciences, Institute of Automation (shuangyi.wang@ia.ac.cn, zengguang.hou@ia.ac.cn).

K. Wang is with the Beijing Advanced Innovation Center for Biomedical Engineering, Beihang University, China (wangkehao1003@gmail.com)

H. Liu is with the School of Biomedical Engineering and Imaging Sciences, King's College London, UK.

Corresponding author: shuangyi.wang@ia.ac.cn

the design concept, the implementation method, the simulation study, and the preliminary phantom and animal tests of this new robot.

## II. DESIGN AND IMPLEMENTATION OF THE ROBOT

### A. Design Concepts

The proposed robot includes an active 2-degree of freedom (DOF) end-effector for actuating the swab and a generic 6-DOF passive arm for the global positioning. The schematic drawing of the end-effector is shown in Figure 2. Within the supporting case, a leadscrew driven linear stage actuated by a stepper motor was mounted. A small geared stepper motor is attached to the front end of the linear stage, controlling the following rotation link. As illustrated in Figure 2, a specially designed swab gripper is attached to the rotation link with its extruded structure tightly fitted into the grove of the rotation link. The swab can be assembled to the swab gripper with its shaft constrained by the constraint hole and held by the gripping cylinder of the swab gripper. The proposed individual structures allow easy attaching and detaching operations between the swab and the swab gripper that can be performed by either the patient or the on-site health professional, making the collection conveniently and both components recyclable. The overall size of the 2-DOF end-effector is $150(L) \times 60(W) \times 40(D)$ mm. The linear stage, the geared stepper motor, and the rotation link are all enclosed by the supporting case and a cover when the linear stage is at its initial position.

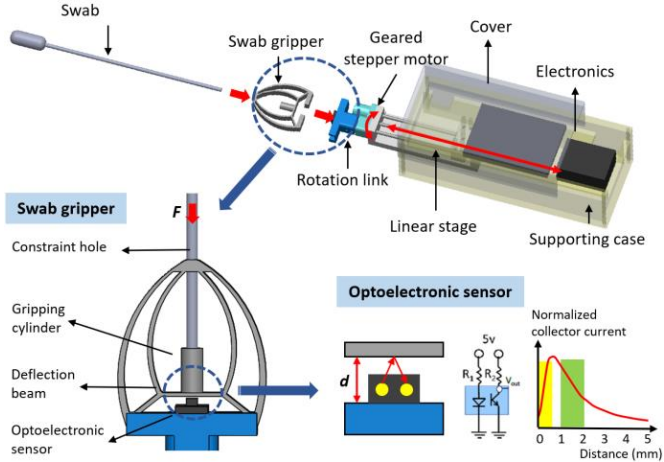

Fig. 2. Schematic illustration of the proposed 2-DOF active end-effector with each component and the mechanism of the force sensing swab gripper shown.

The structure of the swab gripper was designed based on the concept of compliance mechanism. The contact force experienced by the swab when it is in contact with the nasal cavity would result in the largest vertical displacement segment of the structure at the center of the deflection beam. As shown in Figure 2, the displacement is detected by an optoelectronic sensor (QRE1113, Fairchild Semiconductor, California, the United States) which consists of a LED to generate infrared (IR) light and a photo-transistor to receive the reflected light. The measurement circuit is embedded into a breakout board (SparkFun Electronics, Colorado, the United States) to detect the output voltage, which varies depending on the amount of IR light reflected to the sensor with changes of the distance between the sensor and the deflection beam. Two parts of the characteristic curve (as illustrated in Figure 2) show linear relationships between the measured distance ($d$) and the output current. To guarantee a high linearity and sensitivity, the yellow area (measured distance in between 0 to 0.50 mm) in Figure 2 was used in this study. The selected sensor has been proved in our previous studies to be able to achieve large output voltage variations without using an amplifier and demonstrates low level of noises [9, 10]. The overall size of the swab gripper is $38(L) \times 33(W) \times 5(D)$ mm. An additional feature of this design involves an add-on headset that can be attached to a regular chair, used to set the angle of the patient's head, at approximate 70°. The headset has a tilting joint that can be manually adjusted (shown in Figure 3).

### B. Hardware and Software Implementation

As an example used in this research, we employed a 6-DOF passive arm for global positioning and insertion angle adjustments (Figure 3). The passive arm has a global revolute joint (R1), two parallel mechanisms to adjust backwards-forwards and up-down movements (R2 and R3), and a tilting joint (R4) working together with a ball joint (R5 & R6) to fine-tune the angulation of the end-effector, i.e., the insertion angle of the swab. One clamp at the bottom was included to have the arm attached to a table and another clamp at the end was utilized to hold the active end-effector. The components of the arm are widely used for lamp holders and mobile-phone stands, which can be easily bought from the market and assembled to the customized specification. The lengths of the arms are set at 35 mm for this application. The proposed active end-effector was mainly made from 3D printing using Polylactic (PLA) acid and the swab gripper was 3D printed using Polypropylene (PP) material. PLA is a widely used rigid printing material while PP is a semi-rigid flexible material with excellent fatigue resistance.

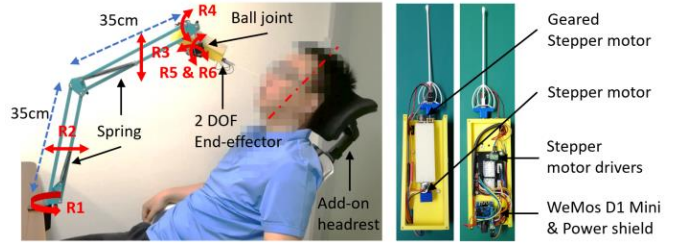

Fig. 3. Implementation of the proposed robot with its passive arm, active end-effector, add-on headrest, and the embedded electronics shown. It should be noted that this photo was taken only for illustration purpose. The in-vivo experiment was not performed in this study.

For electronics, a microcontroller (MCU) WeMos D1 mini with its power shield (core chip based on ESP8266, Espressif Systems, Shanghai, China) and two stepper motor drivers (TCM2209, TRINAMIC Motion Control, Hamburg, Germany) were utilized in the design. The electronics setup is also shown in Figure 3. The whole system is powered by a regular 9V-2A power supply which can be directly plug into the power shield. With the WiFi function of the MCU, we developed a mobile phone-based simple user interface (UI) using the Blynk software (Blynk ver. 2.27.17, Android APP) to remotely control the stepper motors and display the sensor's reading. Joysticks, buttons, and level displays were included into the UI with their functions illustrated in Figure 3. Specifically, with the 2-axis joystick actuated diagonally, the swab can be translated and rotated simultaneously, in both directions of each axis.

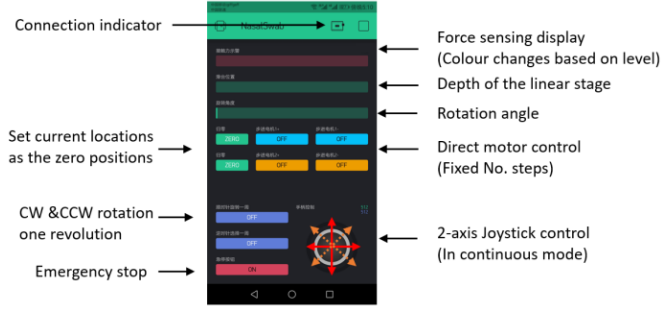

Fig. 4. Screenshot of the robot control software developed in Blynk.

The maximum payload of the robot (measured with a swab mounted and constrained in a tube) is 3.50N and this was measured with a commercial force sensor (JHBM-H3, JNSensor, Anhui, China). The weight of the 2-DOF end-effector with all the electronics included is 0.23kg and the weight of the 6-DOF passive arm is 0.31kg. The manufacturing cost of the system is listed in the table 1 and the total cost price of 400 RMB (~55 USD).

TABLE I
COST PRICE SUMMARY OF THE PROPOSED ROBOT

| Components | COST (RMB) |
| --- | --- |
| Lead-screw driven linear stage | 32 |
| Geared stepper motor for the rotation link | 45 |
| 3D printed supporting case, covers, and rotation link | 120 |
| 3D printed swab gripper | 18 |
| TMC2209 stepper motor drivers and breakout board | 44 |
| WeMos D1 mini and its power shield | 26 |
| QRE1113 optoelectronic sensor | 10 |
| 6-DOF passive arm and two clamps | 40 |
| Add-on headrest with clamp | 45 |
| Regular 9V-2A power supply | 20 |
| *Total costs* | *400* |

## III. FINITE ELEMENT ANALYSIS OF THE SWAB GRIPPER

Finite Element (FE) models of the swab gripper and the swab were developed in ANSYS mechanical APDL (ver. 18.2) using geometric input of IGES file that was created by SOLIDWORKS (ver. 2018). The model is to analyze the displacement performances of the deflection beam of the swab gripper when the swab is subjected to external forces. To justify the design, the expected ranges and magnitudes of the displacements should meet the selected linear range of the optoelectronic sensor. In this study, Young's modulus of 1.70 GPa and Poisson's ratio of 0.43 were used for the swab gripper based on the information provided by the 3D printing manufacturer. Half model was considered as the whole configuration is symmetric about its central plane. Both parts were meshed using quadratic tetrahedral elements (ANSYS element type: SOLID186). Mix u-p element formulation method was chosen to avoid volumetric locking phenomenon given that the selected material has a Poisson's ratio of 0.43 which is close to that of an incompressible material ($v = 0.50$). The swab gripper was meshed using a total number of 32859 elements yielding a total of 55894 nodes and the swab has 4615 elements and 10027 nodes (Figure 5(a)). Contacts between the swab and the constraint hole was modelled as frictionless using Augmented Lagrange algorithm and the contact between the swab and the deflection beam was considered as fully bonded. The boundary conditions were added to constrain the movement of the gripper in Y and Z-axes. 2.50N were applied to the distal end of the swab in seven sub steps as this was the maximum expected force when the swab is in the nasal cavity.

The swab has a thin long shaft and is subjected to buckling under gradually increasing load. However, as this only happens in a limited range and the swab is constrained by the nasal cavity, the deformations do not cause the collapse and will continue to transfer the load. In this study, we focused on the load analysis. The results of the FEA are shown in Figure 5(b). An example color coded displacement profile in the vertical direction and the comparison between undeformed and deformed shapes under 2.50N are shown in Figure 5(b). It can be observed that the maximum displacement occurs at the center point of the deflection beam, resulting in an approximate 0.50mm deformation. The force-displacement curve is shown in Figure 5(c).

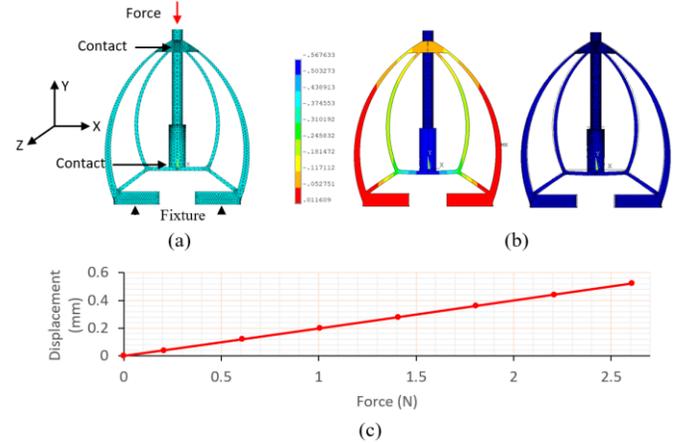

Fig. 5. FE modeling of the swab gripper when subjected to external forces: (a) the mesh results with the force, contacts, and the boundary conditions shown, (b) the displacement results under 2.50N, and (c) the force-displacement relationship with the forces applied from 0 to 2.50N.

## IV. PRELIMINARY EXPERIMENTAL RESULTS

### A. Force Calibration

Force calibration was performed using the commercial force sensor (JHBM-H3, JNSensor, Anhui, China). The setup is shown in Figure 6. To avoid the influence of the buckling effects, a regular daily-used shorter swab with wooden shaft was employed. The linear stage was actuated slowly to generate the target contact forces (from 0 to 3.00N, 0.50N interval) and the according measured output voltages from the optoelectronic sensor were recorded. The loading (blue line) and unloading (black line) curves are shown in Figure 6, where the elastic hysteresis can be observed. As the current application does not require high accuracy when measuring the force, the average values of the measured output voltage during the loading and unloading phases were used for the 2nd order polynomial curve fitting for simplicity purpose and the relationship curve (red line) was programmed into the microcontroller of the robot. Using the red line as the calibration result, the error of the measured samples (N = 12) is 0.24 ± 0.16V (mean ± standard deviation). The expected working range of the sensing gripper is 0 to 2.5N.

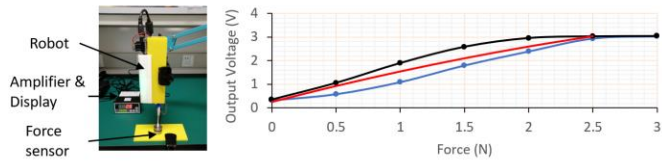

Fig. 6. Setup for the force calibration (left) and the measurement results (right).

### B. Phantom and animal Experiments

To validate the performance of the robot and understand the required force for the NP swab sampling procedure, a commercial nasopharynx phantom (Simon Co., Shanghai, China) and three pig noses (from a food supplier, no ethics required) were utilized, as shown in Figure 7. The phantom has realistic shapes and anatomical structures of the human nose although the silicone material's properties are different to human tissues. The pig noses were obtained from freshly slaughtered pigs (preserved in an ice bag and delivered two days later). Although the anatomical structures of pig nose are different to human nose and the pig noses were also cut to shorter lengths, the tissue's mechanical properties are likely to be close to human. For both experiments, the insertion angles of the swab were set with the swab parallel to the palate as much as possible based on visual observation in the side view. The phantom is stably positioned on the bench while the pig noses required manual supports to remain stable. The robot was actuated from the mobile app with the insertion first and then the retraction combining with the rotation. Specifically for experimental purpose, the measured forces were streamed to a PC with USB cable in real time. For each test, the insertion-retraction procedure was repeated for three times continuously.

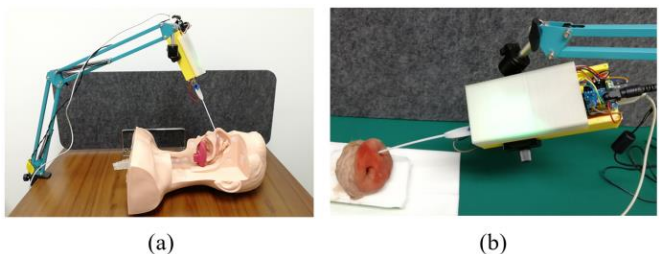

Fig. 7. Experimental setups for (a) the phantom and (b) the pig nose tests.

Upon successful completions of both the phantom and the pig nose experiments, the measured forces for the phantom are shown in Figure 8(a) and an example measurement for the big nose is shown in Figure 8(b). Unfortunately, both the structures of the simulation phantom and the cut pig noses do not allow swab to truly touch the nasopharynx, which would have resulted in another peak force that is currently not presented on each measurement curve. The current peaks arose from the contacts between the swab and the turbinates. By extracting the data of the insertion-retraction procedures, the average force was found to be 0.35 ± 0.10N (mean ± standard deviation) and 0.85 ± 0.64N (mean ± standard deviation) while the maximum force was found to be 0.87 and 2.45 N for the phantom and pig noses, respectively. The maximum identified forces were within the payload and the detectable range of the robot.

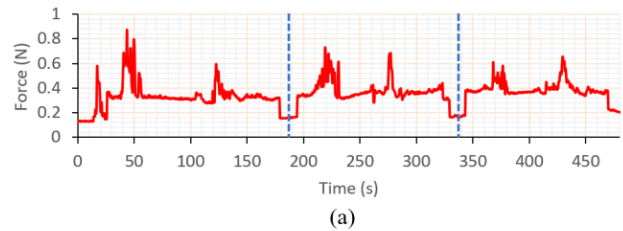

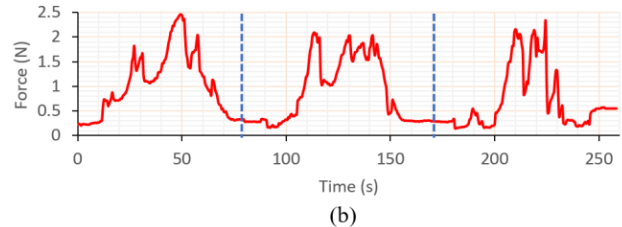

Fig. 8. Force measurement results of (a) the phantom and (b) the pig nose tests.

## V. DISCUSSIONS AND CONCLUSION

This paper introduces a low-cost miniature robot to assist NP sampling. The robot can be made easily using 3D printing technique together with other easy-to-find commonly used components on the market. The robot is intended to be used in a simple way by assembling different pieces together and controlling from a mobile app. The cost price for developing the robot is only 55 USD and can be further reduced if manufactured in great numbers (estimated to be less than 30 USD). The payload of the robot, when controlling a swab with a long thin shaft, is strictly limited by using small motors. Above the threshold, the motor would slip. In addition to the motion system, the robot has a detachable swab gripper which incorporates the force sensing capability based on a simple compliant mechanism and a low-cost optoelectronic sensor. This piece can be easily assembled to and removed from the robot, making it possible for the suspected patient to do self-assembly and disassembly pre and post being tested. The detect range of 0-2.50N of the sensing gripper was justified using the FE analysis. The calibration result indicated that the displacement range (0-0.50mm) of the deflection beam meets the prediction by the FE model and falls into the expected linear range of the optoelectronic sensor. The sensor's design concept was specifically for this low-cost application.

Using the current available resources, we have tested this robot with a commercial nasopharynx phantom and three pig noses. The successful working of the robot was verified and the amounts of forces during the procedure have been preliminarily quantified. Based on the current evidences, it is concluded that the maximum forces were below the payload of the robot and within the detectable range of the sensing gripper. It is also observed that the initial insertion angle of the robot is crucial to be parallel to the palate for the success of the collection, as this would ensure the swab can go into the lower nasal passages. This may require another tilting DOF of the robot to assist the initial angulation alignment or a separate side view monitoring camera system. The current preliminary study was limited by only using the simulation phantom and pig noses, as both cannot fully represent the real nasal cavity of human. Our next step will focus on finalizing the design and pursue ethical approval for in-vivo tests.